# TempOpt - Unsupervised Alarm Relation Learning for Telecommunication Networks

Sathiyanaryanan Sampath
Global AI Accelerator
Ericsson, India
sathiyanarayanan.sampath@ericsson.com

Pratyush Uppuluri
Global AI Accelerator
Ericsson, India
pratyush.kiran.uppuluri@ericsson.com

Thirumaran Ekambaram
Global AI Accelerator
Ericsson, India
thirumaran.ekambaram@ericsson.com

*Abstract*—In a telecommunications network, fault alarms generated by network nodes are monitored in a Network Operations Centre (NOC) to ensure network availability and continuous network operations. The monitoring process comprises of tasks such as active alarms analysis, root alarm identification, and resolution of the underlying problem. Each network node potentially can generate alarms of different types, while nodes can be from multiple vendors, a network can have hundreds of nodes thus resulting in an enormous volume of alarms at any time. Since network nodes are inter-connected, a single fault in the network would trigger multiple sequences of alarms across a variety of nodes and from a monitoring point of view, it is a challenging task for a NOC engineer to be aware of relations between the various alarms, when trying to identify, for example, a root alarm on which an action needs to be taken. To effectively identify root alarms, it is essential to learn relation among the alarms for accurate and faster resolution. In this work we propose a novel unsupervised alarm relation learning technique Temporal Optimization (TempOpt) that is practical and overcomes the limitations of an existing class of alarm relational learning method-temporal dependency methods. Experiments have been carried on real-world network datasets, that demonstrate the improved quality of alarm relations learned by TempOpt as compared to temporal dependency method.

## I. Introduction

Network Operations Centre (NOC) involves monitoring, managing, and maintaining the telecommunications network infrastructure. The NOC ensure its availability, reliability, and performance. Faults in network could lead to service disruptions, nodes in the telecommunication network are designed to generate fault alarms when the state of the node deviates from the normal operational state. Single fault in the network could impact multiple connected nodes and each node could generate more than one alarm as a response. Oftentimes, faults might display transient behaviour (Transient faults), where they spontaneously resolve within a brief period without requiring external corrective interventions. Transient faults and multiple alarms per fault lead to huge volume of alarms.

Various alarm types in a network, such as "Link Down" or "Heartbeat Failure," produce alarm instances—specific occurrences at a given time, often noted with attributes like location. For simplicity, the term "alarm" refers to these types, while "alarm instance" denotes their occurrence. Alarm relation learning helps identify relationships among alarms based on temporal, spatial, and other characteristics drawn from historical data.

In the context of alarm analysis, three prominent scenarios may be considered for learning relationships among alarms,

1. Alarm occurrence; alarms appearing within a very short interval of time, a single fault could lead to burst of alarms, so alarms in the burst are related to one other.

2. Alarm sequence; a fault could propagate within a network, and they could lead to specific sequence of alarms

3. Temporal dependency: a pair of alarms are related to one other if the time interval between the corresponding alarm instances is consistent, this scenario among three where more reliable relations can be established.

NOC engineers handle large volumes of alarms to identify the root causes of faults, often relying on heuristic-based rules derived from experience. However, these manual approaches struggle to scale with increasing network complexity and may miss novel faults as expert-defined rules become obsolete with changes in network characteristics and topology.

Understanding alarm relationships is essential for identifying root causes of network faults. By grouping active alarms into "incidents"—abstract representations of network fault states—NOC engineers can reduce raw alarm data and focus on key patterns. As in [1], while the authors do not explicitly refer to this abstract representation as 'incident,' but it is evident that the prominence of a group of alarms, along with other characteristics, plays a key role in root cause analysis and the creation of Trouble tickets to notify and resolve network faults. Leveraging these incidents empowers NOC engineers to shift their focus from individual incoming alarms to cohesive alarm groupings, ultimately curtailing the mean time required to rectify network faults and preempting service disruptions. Incidents are entirely reliant upon alarm relationships that are robust and dependable. In our work we propose a novel technique of Alarm relation learning that is both scalable and suitable for practical deployments, with improved quality of relations as compared to existing methods that are used in practice, which rely on weak assumptions may not be applicable in Telecommunication networks context

This work focuses on a method for grouping alarms, which will support downstream tasks such as incident detection and root cause analysis—critical for troubleshooting in telecommunications networks. Alarms generally occur as event sequences, and most techniques that learn the structure and grouping of these sequences draw from causal discovery methods, with Granger causality being widely used. In practical cases, methods such as temporal dependency are often employed due to simplicity and effectiveness. Some of these methods are reviewed in Section II. Our main contribution is an improved method over temporal dependency, which is particularly effective in telecommunications networks. We have conducted experiments and field trials to assess the quality of alarm grouping, demonstrating that our method is well-suited for learning the structure and grouping of event sequences, while method works effectively in Telecommunications it can be employed for other event sequences.

## II. RELATED WORK

In this section, we will briefly outline a broad class of methods suitable for causal structure learning and event sequence grouping, with a particular focus on telecommunications networks

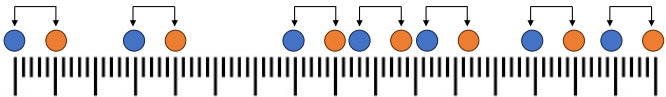

Fig. 1. Blue dots represent instances of independent alarms and orange dots are instance of dependent alarms, they are laid down on the time scale based on its occurrence time. The arrows that connect blue and orange dot is the paring of instances. A simplified scenario illustrating a perfect temporal relation where groups of instnaces are well sepearted

### A. Temporal Dependency based methods

Methods grounded in temporal dependency [2] leverage the consistent reoccurrence of temporal alarm co-occurrences, As illustrated in Fig. 1 An independent alarm A is assumed to cause alarm B, if time interval between various instances of alarm A and alarm B is consistent with some deviation. In practical situations, due to the presence of noise and errors, the temporal gaps between alarm instances of two alarm types A,B may not exhibit perfect uniformity, to address this, the time interval between successive alarm instances is treated as a stochastic variable as mentioned in [2]. This approach generates the most comprehensive type of relationship between alarms: a directed graph that incorporates edge attributes such as the strength of the connection and the time interval ($\delta$) between independent and dependent alarms. The simplicity of adjusting these relationships (through the manipulation of selected thresholds and distribution parameters) and its capability to effectively learn from substantial volumes of data contribute to its widespread adoption and application. However, a significant disadvantage of this method is that the time interval ($\delta$) critical in grouping alarms is arbitrarily chosen, leading to poor performance in telecommunications networks without careful selection, often requiring trial and error. Additionally, the method lacks the ability to incorporate the topological aspects of the network. Our approach seeks to address these deficiencies

### B. Point process based methods

Point process-based methods concentrate on modelling the event generation process itself while simultaneously acquiring insights into relationships. One prevalent example is the utilization of Multivariate Hawkes processes [4], which effectively models various types of event sequences—alarms being one such type. Conversely, Hawkes processes possess a robust theoretical foundation tied to Granger causality [5] and are commonly employed in modelling event sequences to capture causal structures. However, their applicability in the realm of telecommunications networks is limited due to certain assumptions. For instance, the assumption that event sequences are independent and identically distributed renders them unsuitable for real-world data, particularly within the context of telecommunications networks. This is especially true considering that a fault in one node can trigger alarms in different nodes due to network connectivity, thereby violating the a forementioned assumption. Although enhancements like THP [3] have been developed to address this assumption and improve upon to leverage interconnectedness of nodes, these models display increased complexity and require meticulous parameter calibration in comparison to techniques founded upon Temporal Dependency.

### C. Other methods

Various alternative approaches harness the potential of graph structure learning to extract patterns from historical alarm instance data. These methodologies draw upon techniques found in Graph Neural Networks [6]. However, their successful application necessitates the preprocessing of data to align with input structure. Alternatively, certain methodologies make use of hybrid of probabilistic graph models and statistical models such as Hawkes Process to learn the graph structure among alarms [7]. Nevertheless, when it comes to practical implementation, these methods encounter challenges. Specifically, the inherent complexity of the models themselves and their limited adaptability in accommodating domain expertise during the derivation of relationships pose significant hurdles, (based on few experimental validations with domain experts within Ericsson) Incorporating domain-specific considerations into the modeling process proves to be a formidable task due to the intricate nature underlying theoretical aspects thus not presenting a viable alternative to temporal dependency-based methods in practice. All the methods mentioned above are suitable for the task of learning the structure among alarms. However, despite some of them being theoretically robust, they have certain disadvantages, such as the need for parameter tuning, reliance on topology for accurate results, and some suffer due to poor performance in telecommunications settings (based on our experiments), we have developed a method that is practical, suitable for real world deployment

## III. TEMPOPT NOVEL TEMPORAL DEPENDENCY BASED METHOD

In temporal dependency-based methods the basic idea to form relation between alarm A (Independent) and alarm B (dependent) is based on ability to pair as many instances as possible of alarm A and alarm B under a threshold $\varepsilon$, if sufficient number of instances of two alarms A and B tend to exhibit a temporal proximity (similarity in terms of closeness in time) allowing some variability, it's possible establish a dependency Fig. 1 illustrates simplified scenario illustrating a perfect temporal relation. $A_{t'}$ is one such instance alarm of A at $t'$, while $B_T$ is instance of alarm at time T, the idea of pairing, $A_{t'}, B_T$ is valid if Eq.(1) holds true, given a time interval $\delta$ within a maximal deviation $\varepsilon$, applying conditions on number of such pairs of alarm instances of A, B determines the relation $R_{A \rightarrow B}$ and strength; A in this case is independent alarm and B is dependent alarm, condition on number of such pairs is described in Eq.(5) and Eq.(6) in [2] .

Given the simplicity in the formulation and applying condition on number of alarm instance pairs, methods rooted in temporal dependencies provide a straightforward means of managing relations by relying on uncomplicated metrics, including the count of instance pairs and specific parameters such as $\delta$ and $\varepsilon$. The elegance of employing merely two major parameters and simple statistics renders the models easily tunable and applicable in real-world scenarios. This simplicity also facilitates the incorporation of domain expertise, a particularly critical factor within telecommunications networks. Through the manipulation of these parameter pairs pertaining to alarm instances A and B, the domain-specific knowledge can be effectively integrated into the model.

$$|T - \delta - t'| < \varepsilon \text{ and } T > t' \qquad (1)$$

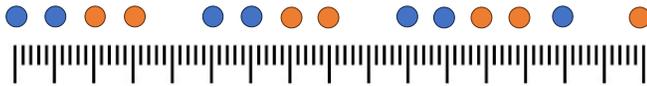

Fig. 2. Perfectly spaced alarms but overlapping pairs where a orange dot (dependent alarm instance) can be paired with any of two blue dots (independent alarm instance)

*A. Drawbacks of Temporal Dependency based methods*

Fig. 1 demonstrates an ideal scenario, where independent and dependent pairs of alarm instances are well separated, a one to one relation is established between orange and blue dot as there is no overlap between various pairs of orange and blue dots. The concept of considering the time interval ($\delta$) between consecutive alarm instances as a stochastic variable as described in [2] allows for the accommodation of random noise in the temporal spacing between alarm instances of A and B across different pairs, this notion thus avoids the restriction of time-gap / spacing between instances across pairs to be uniform or fixed. The variation in the time interval ($\delta$) can be handled by bounding with a tolerance threshold $\varepsilon$ as shown in Eq.(1), there is a critical problem that commonly occurs in real time telecommunications data that is illustrated in Fig. 2

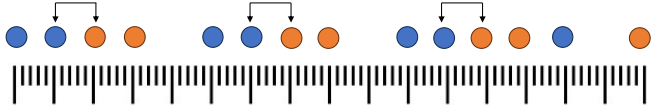

Fig. 3. Blue dots are instance of independent alarms and orange dots are instance of dependent alarms, they are laid down on the time scale based on its occurrence time. The arrows that connect blue and orange dot is the paring created by current solutions, they look at the immediate next occurrence, under many scenarios this could lead to incorrect findings.

As illustrated in Fig. 2 for a given sequence of alarms (instances of two different alarms), identifying instance of dependent alarm that must be paired with instance of independent alarm is a combinatorial problem due to the overlap between instances, as every instance of independent alarm has potential to be paired with another instance of dependent alarm that occurred after it, this overlap is regularly observed in telecommunications network, the well separated scenario illustrated in Fig. 1 is a rare phenomenon. One simple way to address this issue of overlapping pairs is to consider immediate next occurrence of dependent alarm instance after an independent alarm instance as its pair, this is illustrated in Fig. 3, but this assumption could lead to inaccurate results because of complex pattern of occurrences of independent alarm instances, also in Fig. 3 it can be observed many instances do not get paired due to this weak assumption, thus effecting the ability to determine the alarm relations effectively.

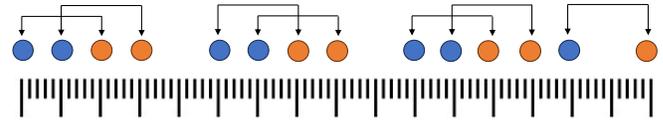

Fig. 4. Alternate way to create pairs for the same given sequence of alarms.

Fig. 4 illustrates an alternative method for creating pairs, a strategy can be envisioned in the form of a combinatorial task, where each instance of an independent alarm has the potential to form a pair with every subsequent instance of a dependent alarm, illustrated in Fig. 5 The objective here is to optimize the pairing of instances. Nonetheless, this approach also requires to determine optimal value $\delta$(time gap between instances), which further tackles the issue of identifying the most suitable $\delta$ for establishing a relationship between alarms A and B.

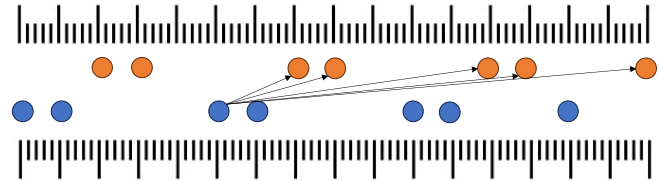

Fig. 5. Every instance of independent alarm paired with every instance of dependent alarm

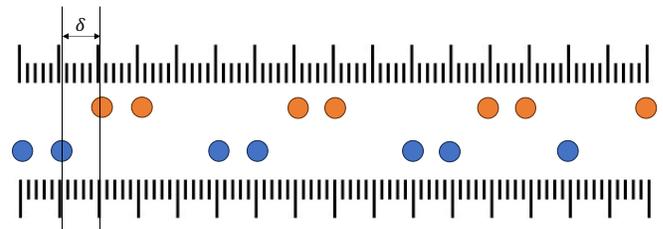

Fig. 6. $\delta$ is the time offset made to independent alarm to create a matching, during optimization $\delta$ is chosen such a way that number of pairs observed in the sequence is maximized. The original gap among instances of both independent and dependent alarms is maintained.

*B. Our Solution*

$$\max_{\delta} \text{Number of Alarm Pairs} \qquad (2)$$

To address the issue of combinatorial explosion of determining pairs of instances that could help establish a relation between an independent and dependent alarm, the problem could be modeled as optimization problem. While preserving the temporal intervals/gap between occurrences of instances of both independent and dependent alarms, the independent alarm instances undergo a $\delta$-based offset to maximize the formation of pairs as in Eq. (2), $\delta$ is the time interval between alarms, illustrated in Fig. 5. Under discrete condition an independent and dependent alarm instance can be called as pair, if post offset, they are spaced less than $\varepsilon$. $\varepsilon$ is a very small tolerant value.

## C. Details of optimization formulation

The problem is modeled as continuous nonlinear optimization problem where core part is defining the paring condition. The discrete condition described in section B cannot capture the continuous nature of variability in the time gap, we need a continuous function to represent the pairing. We define continuous pairing function that can take of the form as shown in Eq.(3), where s is horizontal stretching factor and EP is pairing error as in Eq.(4), function should take a value of 1 for absolute pair, when paring error is zero and quadratically the score should decay when the error increases and takes a value of zero at infinite paring error as shown in Fig. 7. Paring Error as in Eq.(4), is a function of $\delta$ where as $b_j$ and $a_i$ are constant for a given pair of alarm instances, they hold occurrence time of each of instances respectively.

Next to address the possible combinations of pairing, every independent alarm instance is potentially paired with subsequent occurrences of all dependent alarm instances as shown in Fig. 5. The specific form of optimization is described in Eq.(5).

$$Pairing\ Function = \frac{1}{(1 + sE_P^2)} \quad (3)$$

$$Pairing\ Error(E_P) = (b_j - a_i - \delta) \quad (4)$$

$$\max_{\delta} \sum_{c_x \in C} \sum_{i} \sum_{j,\, a_{icx} < (b_{jcx}+\tau)} \frac{1}{(1 + s(b_{jcx} - a_{icx} - \delta)^2)} \quad (5)$$

Subjected $to\ 0 \leq \delta \leq\ Max\delta$

$c_x \in C$ is the $x^{th}$ network context, $a_{icx}$ $i^{th}$ occurrence time of the alarm A at context $c_x$, $b_{jcx}$ $j^{th}$ occurance time of the alarm B at context $c_x$, $\tau$ is a small positive constant.

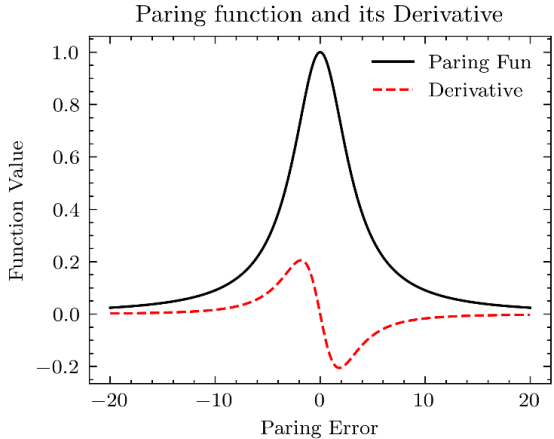

Fig. 7. Plot of the continuous paring function and its derivative, this captures the specfic form of pairing function

To account for delay in alarm generation and reporting, a small positive integer $\tau$ is introduced in the expression. Additional context variables are also introduced to accommodate spatial relatedness of alarm instances, since network nodes are connected across, and information can be made available in the form of network topology. Alarms are related only within a context say alarms occurring within a node/site or with in a connected component of network; a set of nodes/sites. This additional context inclusion is natural to our formulation which is difficult in traditional temporal dependency methods which rely purely on temporal relatedness. As previous temporal dependency methods were not specifically designed with view of telecommunication networks, it was sufficient in those methods to consider only temporal relationship without the need of the spatial characteristics that exits in a telecommunication network, due to the connectedness.

**Algorithm 1:** Algorithm to compute alarm relation graph with Temporal Optimization

**Require:** Alarm set A = {A1, A2, A3, …}
**Ensure:** alarm graph G
1   Alarm graph candidate set C = Permutation (A, 2)
2   for c in C do
3       r <- relation metrics for the pair (c.source, c.target)
4       if r > desired value
5           add edge (c.source, c.target) in graph G
6       end if
7   end for

## D. Optimization Strategy

As shown in the Fig. 8, this is a single dimension optimization problem, it is easy to visualize the objective function with respect to the optimization variable. Objective function is highly nonconvex and have multiple local maximums. To achieve the global maximum, optimization can be initiated multiple times using random initial values for $\delta$. Alternatively, a grid search for $\delta$ can be conducted using a specified number of disjoint bins, such as 0 to 1000, 1001 to 2000, and so forth. To enhance convergence and minimize brute-force repetitions, the following strategy is proposed

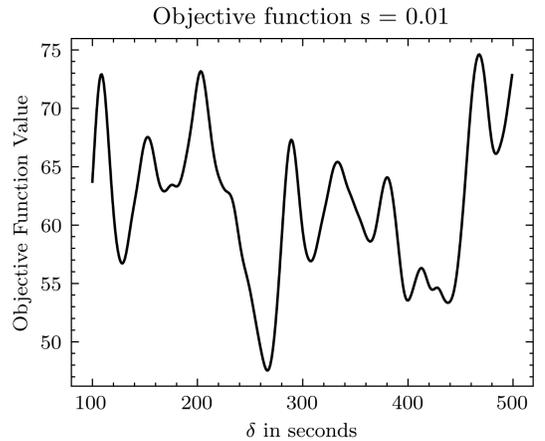

Fig. 8. Plot of objective function for a given s, that shows many local minima and maxima

Perform a grid search with random initial values selected from predetermined disjoint bins. These will likely converge to a few local maxima. Using the results obtained as the initial values, perform the optimization by increasing the value of s by a factor of 10, for example, from 1e-5 to 1e-4. Repeat this process, each time increasing s by a factor of 10, until s equals 1. The final optimal result is the one that yields the maximum objective function. If multiple peaks are found with only slight differences in the objective value, it is advisable to choose the one with the smallest $\delta$ value. The behavior of the objective function for different values of s can be observed in Fig. 9.

**Algorithm 2:** Algorithm to compute relation metrics for a given pair of source & target alarm

**Require:** Alarm pair (source, target), context c = {c1, c2, …} instance occurrence time data for a given context {s1cx, s2cx, …} {t1cx, t2cx, …}
**Ensure:** relation metrics
1  for cx in C do
2      Pcx <- Potential pair for context cx
3  end for
4  P <- Concat (Pcx) Potential pairs at each context
5  $\delta$ <- find delta that maximize the paring as given in Equation 5
6  Compute paring metrics, $S_{ratio}$, $T_{ratio}$, Quality

### E. Alarm Relation Graph

The $\delta$ obtained via the optimization is the time interval of alarm instance pair of A, B, across the majority of pairs of instances that sufficiently establishes a dependency relation $R_{A \rightarrow B}$. Support is number of instances of independent alarms; confidence is Number of Pairs/Support. Using the algorithm 1 and 2, we would be able to obtain the various relations among different alarms observed in the historical data and generated metrics will help to prioritize the most important relations. Sample relations are also qualitatively validated with domain experts. Due to confidentiality, we are unable to present the actual alarm relationships obtained. However, to provide perspective, we have defined several metrics to assess the quality of the learned alarms and to compare them against the Temporal Dependency method. While we acknowledge that benchmarking datasets would ideally be used for comparison, in this case, we only provide comparisons based on the defined metrics

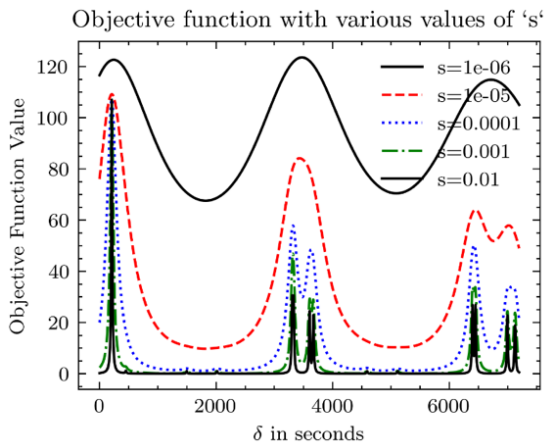

Fig. 9. Plot of objective function for various values of s

### F. Implemenation

We collected various alarm logs from a Network Operations Center (NOC) using real data. Our procedure analyzes pairs of alarms at a time to find the optimal $\delta$ and determine the strength of their relationship. Since each pair is processed independently (e.g., alarms A and B), the implementation can be easily parallelized for different combinations

The metrics for evaluation could be computed with data aggregation functions, they could be used as initial filters to identify candidate pairs for optimization algorithm to expedite the search, the search is implemented with Python library SciPy's optimize module. [8]

## IV. EXPERIMENTS AND RESULTS

### A. Data

Alarm logs collected it has following fields, ID: unique identification of each instance of alarm, Node ID: Network element that generated the alarm, Site ID: location where the network node is present, Node Type: Type of the Network node, Alarm: X733 Specific Problem (Alarm Type or Alarm), First Occurrence: Time stamp of the alarm. Depending on the use case either Site or Node could be used as alarm relation scope or context. The data is collected from a real Operator from the period March 1st, 2023, to March 30th, 2023, from 2000+ telecom sites which have a total of 1232 alarm types.

### B. Metrics

Due to the challenges in presenting actual alarm relationships and the absence of benchmark datasets, we have defined various metrics to compare the quality of alarm relations learned using our Temporal Optimization (TempOpt) method against those obtained with the Temporal Dependency method. As causality-based methods are effective when considering topology and meticulous fine tuning of hyperparameters, since we are comparing these methods in a setting without topology, Temporal Dependency serves as the best comparison method. Our method allows for the incorporation of topology, as described in equation (5), by using context variables that capture spatial relationships through network topology. However, for the datasets in this experiment, topology information is unavailable, preventing us from benchmarking our results against algorithms that include topology. In future experiments, with datasets that include topology, we will be able to benchmark our algorithm against other topology-dependent algorithms, such as THP [5]

These metrics offer insights into the quality of alarm relationships in the absence of ground truth—a common challenge in Telecommunications. Additionally, due to confidentiality issues, we are unable to provide qualitative validation from experts. Nevertheless, we believe these metrics are indicative of our method's performance. We use these metrics to compare our method with the temporal dependency-based approach defined in [2], where we implemented our version of the method described in that reference. Generally, a higher number of relationships across different metrics suggests better quality in the learned alarm relations. This conclusion is

based on empirical observations of learned alarm relations validated by experts across various NOCs

$$S_{ratio} = \frac{dis\_match}{count\_s} \quad (6)$$

$$T_{ratio} = \frac{dis\_match}{count\_t} \quad (7)$$

$$\text{Quality} = S_{ratio} * Min(T_{ratio}, e^{-(T_{ratio}-1)}) \quad (8)$$

*dis_match* is the discrete match of source and target, *count_s* is the total instances of source alarm in a context. $S_{ratio}$ or Source Ratio as in Eq.(6), is the ratio between combination of source/independent and target/dependent occurrence and source alarm occurred in each context. $T_{ratio}$ or Target Ratio as in Eq.(7), is ratio of source, target combination and total instances of target, It should be noted that this ratio could be greater than 1, as multiple source alarms could be paired with single target alarm. Quality as defined in Eq.(8), for a relation between a combination of alarms, is stronger if target alarm instance occurs for every instance of source alarm and target doesn't occur without source, this ideal condition would be met when both $S_{ratio}$ and $T_{ratio}$ is 1 or in practical scenario with noise, the product of both should be higher. Quality is defined as product of $S_{ratio}$ and $T_{ratio}$ post compensating $T_{ratio}$ for values greater than 1.

*Table I Results*

| Data for 30 Days, 1 batch each day | | |
|---|---|---|
| Table column subhead | Temporal Dependency | Temporal Optimization |
| No. Edges; Th Discrete matches > 20 | 1388 | 25790 |
| No. Edges; Discrete matches > 20 & $S_{ratio}$ > 0.2 | 830 | 14596 |
| No. Edges; Discrete matches > 20 & $S_{ratio}$ > 0.2 & Quality > 0.2 | 83 | 3962 |
| No. Edges; Discrete matches > 20 & Time > 600 | 25 | 20882 |

The results described in Table I are a comparison drawn between our implementation of Temporal dependency described in [2] ( not exact but closely aligns to formulation) and our method of Temporal Optimization on real operator data, we demonstrate considerable improvement in quality of relations based on metrics described in the metrics section. We first clearly observe Our Temporal Optimization method produces significantly more relationships under various conditions, as shown in the Table I. In telecommunications, missing relations (false negatives) are a major challenge, but while a higher number of alarm relations may increase false positives, these can be filtered using simple statistical methods. We emphasize that our method's metrics are designed to assess the non-randomness in the occurrence of alarm pairs, aiming to identify those that occur together as frequently as possible thus effectively getting better relations among alarms.

## V. CONCLUSION

We propose Temporal Optimization, a novel method for learning alarm relationships in telecommunications networks, which shows potential for decent results and is suitable for practical deployment. Empirical observations and qualitative assessments suggest that it produces better alarm relations traditional temporal dependency methods. Our approach addresses the limitations of existing temporal dependency methods, which, despite their simplicity and popularity, may not always be effective in practice. We conducted experiments using real-world telecommunications operator data and validated the results with domain experts to evaluate the quality of the learned relationships. Although our experiments and validations were based on real operator data rather than benchmark datasets, further detail and convincing evidence could improve the presentation of these findings.

Additionally, in future we plan to collect benchmark datasets from sources such as [1]. Currently, we have the capability to learn cross-domain relations with the context inclusion described in Eq. (5). Following these developments, we aim to benchmark our results against cross-domain alarm relation learning algorithms, such as THP [2] and substantiate our method

ACKNOWLEDGMENT

Authors sincerely thank Jörg Niemöller Wenfeng Hu, Erik Sanders and Thea Emilsson from Ericsson CNS and Marios Daoutis from Ericsson Research for their valuable inputs and suggestions.